# LLM-Based Evaluation of Low-Resource Machine Translation: A Reference-less Dialect Guided Approach with a Refined Sylheti-English Benchmark

Md. Atiqur Rahman[1], Sabrina Islam[1] and Mushfiqul Haque Omi[2]

[1]Islamic University of Technology, Gazipur, Bangladesh
[2]United International University, Dhaka, Bangladesh
atiqurrahman23@iut-dhaka.edu
sabrinaislam22@iut-dhaka.edu
mushfiqul@cse.uiu.ac.bd

**Abstract.** Evaluating machine translation (MT) for low-resource languages poses a persistent challenge, primarily due to the limited availability of high-quality reference translations. This issue is further exacerbated in languages with multiple dialects, where linguistic diversity and data scarcity hinder robust evaluation. Large Language Models (LLMs) present a promising solution through reference-free evaluation techniques; however, their effectiveness diminishes in the absence of dialect-specific context and tailored guidance. In this work, we propose a comprehensive framework that enhances LLM-based MT evaluation using a dialect guided approach. We extend the ONUBAD dataset by incorporating Sylheti-English sentence pairs, corresponding machine-translations, and Direct Assessment (DA) scores annotated by native speakers. To address the vocabulary gap, we augment the tokenizer vocabulary with dialect-specific terms. We further introduce a regression head to enable scalar score prediction and design a dialect-guided (DG) prompting strategy. Our evaluation across multiple LLMs shows that the proposed pipeline consistently outperforms existing methods, achieving the highest gain of +0.1083 in Spearman correlation, along with improvements across other evaluation settings. The dataset and the code are available at **https://github.com/180041123-Atiq/MTEonLowResourceLanguage.**

**Keywords:** Machine Translation Evaluation, Large Language Models (LLMs), Low-Resource Languages, Prompt Engineering.

## 1 Introduction

Evaluating the quality of machine-translated text traditionally relies on regression systems that depend heavily on human-generated reference translations [1]. However, producing such references is costly and difficult to scale, especially for low-resource, dialect-rich languages. These challenges are further intensified by the scarcity, inconsistency, or complete absence of annotated ground truth data [2]. As a result, the lack of scalable evaluation frameworks not only hampers the advancement of these lan-guages in digital space but also threatens their long-term preservation, including the rich cultural and historical heritage they embody.

With the emergence of Large Language Models (LLMs), new opportunities have surfaced for reference-less evaluation of machine translations. LLMs are capable of modeling semantic relationships and contextual meaning across languages, making them promising tools for assessing translation quality [3]. However, generating reliable quality scores such as Direct Assessment (DA) scores [4] remain a non-trivial task. These scores demand sensitivity to both grammatical accuracy and cross-lingual semantic equivalence, which LLMs may not consistently deliver without targeted guidance.

In this work, we investigate the effectiveness of LLMs for machine translation (MT) evaluation in low-resource, dialect-rich settings, where traditional methods falter due to the absence of reference translations and annotated data. We propose a reference-less evaluation framework that leverages LLMs' ability to interpret regional language variation, guided by a dialect specific prompting strategy that enables inference of DA scores without full finetuning. Our key contributions are:

- We expanded an existing dataset by adding 520 new Sylheti source sentences, their human-translated English references, machine-generated translations for the entire set of 1500 sentences, and Direct Assessment (DA) scores rated by native Sylheti speakers.
- Dialect Guided prompting strategy to improve the reference-less MT evaluation of low resource language.
- We enhance the model by incorporating a dialect-aware tokenizer and a lightweight regression head that leverages the rich feature representations of the pre-trained LLM to predict Direct Assessment (DA) scores.

## 2   Background

The WMT 2023 shared task [5] marked a significant advancement in low resource MT evaluation by introducing new language pairs and providing diverse evaluation resources. A key challenge addressed was quality estimation (QE) without reference translations, promoting scalable evaluation methods. However, dialect-rich language varieties remained largely overlooked. In many South and East Asian countries, distinct dialects are spoken across various regions and communities. These dialects often suffer from limited linguistic resources, largely due to their lack of digital presence [6]. One core barrier to their inclusion in the digital sphere is the absence of effective MT evaluation frameworks tailored for dialect rich, low resource languages. Without effective methods to assess translations in these dialects, there is limited incentive for the development of MT systems targeting them. This persistent gap in technological inclusion not only reinforces their digital marginalization but also poses a significant threat to the preservation of the cultural heritage and linguistic identity embedded within these dialectal varieties.

Although XLM-RoBERTa-based quality estimation (QE) models dominated early submissions to the WMT shared tasks [7], the emergence of large language models (LLMs) signals a potential paradigm shift in the evaluation of machine translation (MT). Sindhujan et al. investigated the application of LLMs for reference-less MT evaluation in low resource scenarios[8]. Their approach used Annotation Guided (AG) prompting, providing LLMs with the same guidelines as human DA annotators to instruct quality scoring based on defined criteria. While this strategy yielded promising outcomes, it lacked explicit mechanisms to account for language or dialect-specific features, thereby limiting its effectiveness in linguistically diverse or dialect-rich environments.

General purpose LLMs are typically trained on corpora that underrepresented low resource languages, resulting in tokenizers that poorly segment dialect specific terms. This affects the model's ability to interpret and evaluate translations accurately. Chalkidis et al. fine-tuned BERT for legal text with tokenizer adjustments to better capture legal terminology [9]. Dagan et al. highlights through extensive experiments that extending or replacing a tokenizer can lead to significant improvements [10]. However, in scenarios without broad multilingual training support, building a tokenizer from scratch for machine translation (MT) evaluation may compromise the model's inherent multilingual capabilities, which are equally essential for reliable evaluation across language pairs.

Moreover, LLMs tend to generate verbose outputs, making them unsuitable for scalar DA scoring. To address this, recent research has explored leveraging LLM embeddings as feature representations for downstream regression tasks [11]. These embeddings have shown promising results, often outperforming traditional feature engineering approaches in high-dimensional regression settings. Thus, embedding-based regression presents a viable and effective pathway for integrating their rich representations into structured evaluation tasks.

## 3  Methodology

We propose a comprehensive framework that combines an enhanced Sylheti-English dataset annotated with Direct Assessment (DA) scores, a novel dialect guided prompting strategy, and the integration of a dialect aware extended tokenizer. In addition, we train a regressor head to better align LLM output with human evaluation scores, enabling the generation of consistent and interpretable translation quality assessments, even in the absence of reference translations. The overview of our proposed pipeline is shown in Figure 1.

### 3.1  Sylheti-English Benchmark

Bengali is a low resource language in natural language processing (NLP), with dialects like Sylheti, Chittagong, and Barisal being even more underrepresented. To address this, ONUBAD [12] introduced a parallel corpus translating these dialects into Standard Bangla and English using expert translators, providing 1,540 words, 130 clauses, and 980 sentences per dialect. For our study, we focused on the Sylheti-English pair and adapted the dataset for LLM-based machine translation (MT) evalua-tion. We extracted the 980 Sylheti-English sentence pairs, corrected inconsistencies, and added 520 new sentence pairs, all translated by native speakers and cross-validated for accuracy, resulting in 1,500 high-quality pairs. To simulate a real-world MT evaluation scenario, we generated translations using the NLLB-200 model [13], recognized for its multilingual capabilities [5]. Two native Sylheti speakers evaluated the outputs using Direct Assessment (DA) guidelines [14], scoring based on semantic equivalence and fluency. Scores were averaged and z-normalized to reduce inter annotator variability and outliers [8].

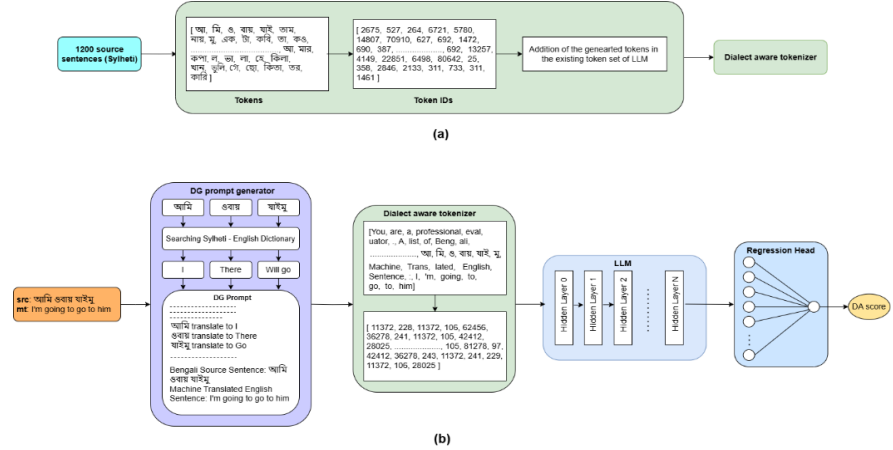

**Fig. 1.** Overview of the proposed pipeline. (a) illustrates the process of creating a dialect-aware tokenizer using dialect-specific vocabulary and (b) presents the complete pipeline, integrating the extended tokenizer with dialect guided prompting and regression-head.

### 3.2 Dialect Guided Prompting Strategy

We designed a novel Dialect Guided (DG) prompting strategy to guide LLMs in the evaluation. The DG prompt, illustrated in Figure 2, explicitly incorporates dialectal context into the model's instruction. The prompt begins with a Bengali source sentence exhibiting Sylheti dialectal features. Each word in the sentence is cross referenced with a curated Sylheti-English dictionary containing dialect specific terms and their English equivalents. When a match is found, the corresponding Sylheti term and its English translation are included as part of the prompt's contextual information, giving the LLM explicit reference points for interpreting dialect-specific vocabulary. The final prompt consists of three parts: a glossary of Sylheti words and their English meanings, original Sylheti-influenced Bengali source sentence, and corresponding machine translated English sentence. We also introduced the Dialect and Annotation Guided (DAG) prompt that extends the DG prompt by incorporating annotation guidelines.

```
You are a professional machine translation evaluator.
You will be given:
        - A list of Bengali words influenced by the Sylheti dialect,
          along with their English translations, to help you better understand the dialect.
        - A sentence in Bengali that includes words from the Sylheti dialect.
        - A machine-translated English sentence.

After evaluating whether the English sentence is an accurate translation of the Bengali
source, your task is to provide only a score from 0 to 100, no need for explanation. The
score may be a floating-point number.

For your reference, here is a list of Bengali words influenced by the Sylheti dialect,
along with their English translations:

আমি translate to I
ওবায় translate to There
যাইমু translate to Go

Now evaluate the following and give only the score:
Bengali Source Sentence: আমি ওবায় যাইমু
Machine Translated English Sentence: I'm going to go to him
Score:
```

**Fig. 2.** DG Prompt

### 3.3 Dialect Aware Tokenizer

Figure 1(a) shows how we adapted the tokenizer associated with LLM to handle Sylheti dialectal variations. We trained a tokenizer on Sylheti sentences and compared its vocabulary with the original LLM tokenizer. Unique Sylheti tokens were added to the LLM's tokenizer, improving its ability to process dialect influenced text and boosting performance in downstream tasks like translation evaluation.

### 3.4 Regression Head

To improve DA score prediction, we added a regression head on top of a frozen LLM. This head was fine-tuned to map the LLM's contextual outputs to human-annotated scores, mimicking expert evaluation. Freezing the LLM preserved its language understanding and minimized overfitting on our small dataset (1200 samples). The head, matching the LLM's output dimension, was trained with MSE loss to align with human scoring. This setup enabled efficient, focused quality estimation without full model fine-tuning.

### 3.5 Implementation Details

We used a Byte-Level BPE tokenizer [15] to capture dialectal variations through subword tokenization. Hugging Face Transformers managed the LLMs, while PyTorch was used to implement and fine-tune a lightweight regression head for predicting DA scores. Input sequence lengths were limited to 512 tokens to balance performance and efficiency. Adam optimizer was used with a 2e-5 learning rate to avoid overfitting on the small dataset. Prompts followed Llama style chat templates for consistency. To optimize memory and speed, we applied 4-bit quantization and fp16 precision, enabling larger models and batch sizes on limited hardware. All experi-ments ran on a single NVIDIA RTX A5000 GPU (24GB VRAM) with 50GB system RAM, motivating design choices that favored efficiency and stability.

## 4 Results

### 4.1 Performance of the Proposed Pipeline

We evaluated the performance of Llama-2-7B [16], Llama-2-13B, OpenChat [17] and Gemma [18] using three different pipelines: the SOTA AG (Annotation Guided) [cite], our proposed pipeline with the DG prompting, and our pipeline with DAG (Dialect + Annotation Guided) prompt instead of the DG prompt. To assess model performance, we computed both the Pearson correlation coefficient [19] and the Spearman rank correlation coefficient [20] between the predicted and actual Direct Assessment (DA) scores.

**Table 1.** Evaluation of proposed pipeline utilizing the Pearson correlation coefficient.

|  | AG | DG | DAG |
| --- | --- | --- | --- |
| Llama-2-7B | 0.1779 ± 0.0023 | **0.2443 ± 0.0015** | 0.2039 ± 0.0086 |
| Llama-2-13B | 0.1765 ± 0.0008 | **0.279 ± 0.0041** | 0.2196 ± 0.0106 |
| OpenChat | 0.2299 ± 0.0144 | **0.2826 ± 0.0091** | 0.2614 ± 0.0187 |
| Gemma | **0.3247 ± 0.0012** | 0.3079 ± 0.0033 | 0.2945 ± 0.0035 |

Table 1 presents the Pearson correlation results across all models and pipelines. Notably, DG consistently outperformed AG and DAG for Llama-2-7B, Llama-2-13B and OpenChat, achieving the highest correlation scores among all evaluated settings. Similarly, Table 2 displays Spearman rank correlations, where DG again resulted in the top performance for the three models.

**Table 2.** Evaluation of proposed pipeline using the Spearman rank correlation coefficient

|  | AG | DG | DAG |
| --- | --- | --- | --- |
| Llama-2-7B | 0.1506 ± 0.0028 | **0.2337 ± 0.0046** | 0.1794 ± 0.0072 |
| Llama-2-13B | 0.1496 ± 0.0018 | **0.2579 ± 0.0064** | 0.1951 ± 0.0121 |
| OpenChat | 0.1984 ± 0.0143 | **0.2852 ± 0.0153** | 0.2441 ± 0.0209 |
| Gemma | **0.3566 ± 0.0021** | 0.3488 ± 0.0023 | 0.3322 ± 0.0014 |

While Gemma showed slightly better performance under AG, AG consistently underperformed across the remaining models. DAG showed improvement over AG but did not surpass DG, confirming the effectiveness of incorporating dialect specific cues into the prompt.

The highest Pearson and Spearman correlations, 0.2826 and 0.2852, respectively, were achieved by OpenChat using our pipeline. DG led to an approximate 58% improvement in the correlation scores for Llama-2-13B compared to AG which is a significant accomplishment.

**4.2 Ablation Study**

To assess the individual impact of each proposed component of our pipeline, specifically the dialect aware tokenizer, the fine-tuned regression head, and the dialect guided prompting strategy, we performed an ablation study using the OpenChat model, which demonstrated the strongest overall performance in our experiments. The results, presented in Table 3, demonstrate the incremental benefits of each component.

**Table 3.** Ablation Analysis of the Proposed Pipeline Using Pearson Correlation Coefficient (OpenChat Model)

| Dialect Aware Tokenizer | Regression Head | DG Prompt | Pearson Correlation Coefficient |
|---|---|---|---|
| × | × | × | -0.1446 |
| ✓ | × | × | 0.0336 |
| ✓ | ✓ | × | 0.2312 |
| ✓ | ✓ | ✓ | **0.2734** |

Without any modifications, OpenChat achieved a baseline Pearson correlation score of –0.1446, indicating poor agreement between predicted and reference DA scores. Introducing the dialect-aware tokenizer resulted in a substantial improvement, increasing the score by approximately 123% to 0.0336 which is a critical shift from negative to positive correlation, indicating initial alignment between predicted and reference scores. Incorporating the fine-tuned regression head further boosted the score to 0.2312, enabling the model to better generate scalar quality predictions. Finally, applying the dialect guided prompting strategy raised the score to 0.2734, underscoring the value of incorporating dialect-specific cues directly into the evaluation process.

## 5 Discussion

The evaluation results in Table 1 underscore the effectiveness and robustness of our proposed methodology in the context of reference-less machine translation (MT) evaluation. Our pipeline mostly outperforms the baseline Annotation Guided (AG) strategy and its combined variant (DAG), establishing the superiority of embedding dialect specific linguistic cues directly into the prompt.

A key insight from the results is that the DG strategy improves the model's performance in all LLMs but also does so more significantly than the DAG setup. While DAG combines both scoring guidelines and dialectal cues, the fact that DG alone surpasses DAG suggests that dialectal awareness contributes more meaningfully to accurate MT evaluation than general annotation instructions. This highlights the strong impact of dialect guided prompting and dialect aware tokenization in capturing subtle semantic and fluency nuances in low-resource settings.

When examining model wise performance, Llama-2-13B saw the most substantial gain under the DG setup, improving from 0.1765 (AG) to 0.279 (DG). This shows that larger models capable of disambiguating dialect-specific expressions are better at leveraging the richer contextual signals offered by dialect-oriented strategies.

Similarly, OpenChat, which is pretrained on a wide range of instruction following and evaluation tasks, showed strong performance even though it wasn't specifically trained for MT evaluation. Our pipeline improved its correlation score as well from 0.2299 (AG) to 0.2826 (DG) showing that dialect awareness can complement instruction tuned models with general purpose evaluation capabilities.

Turning to Gemma, we observed that although it achieved the highest Pearson correlation overall (0.3247 in AG), it did not yield its best performance under our DG prompting (dropping slightly to 0.3079). Though the difference in Pearson correlation between AG and DG for Gemma is marginal, Spearman rank correlation is nearly identical to the best-performing setup. This suggests that our pipeline still preserves ranking fidelity, even if it marginally trails in raw score correlation.

We employed chat-tuned variants of Llama-2and OpenChat in our experiments, but did not include the chat-tuned variant of Gemma. We attribute Gemma's lower performance with DG to its lack of instruction or chat tuning. Unlike Llama-2 and OpenChat, which are fine tuned for instruction rich, multi turn prompts, Gemma lacks native support for such interactions, making our dialect guided prompts a better fit for chat optimized models.

## 6    Conclusion

In summary, our pipeline sets a strong precedent for reference-less machine translation evaluation, particularly in low-resource, dialect-rich settings. By augmenting large language models with explicit dialectal context through tokenizer extension and targeted prompting, we significantly improve alignment with human judgments. The approach proves effective across diverse LLMs, with larger models like LLaMA-2–13B showing the highest gains due to their greater contextual capacity. Even models less suited for multi-turn or dialect-sensitive tasks, such as Gemma, achieve competitive results, underscoring the adaptability and robustness of the proposed framework.